\documentclass[conference]{IEEEtran}
\usepackage{times}

\usepackage[numbers]{natbib}
\usepackage{multicol}
\usepackage[bookmarks=true]{hyperref}

\usepackage{booktabs} 
\usepackage{amsthm}
\usepackage{graphicx}
\usepackage{caption}
\usepackage{subcaption}
\usepackage[linesnumbered,ruled,vlined]{algorithm2e}
\usepackage{algorithmicx}
\usepackage{amsmath}
\theoremstyle{plain}
\usepackage{mathtools}
\usepackage{epsfig}
\usepackage{amssymb}
\usepackage{epstopdf}
\usepackage{multirow}
\usepackage{array}
\usepackage{url}
\usepackage{color}
\usepackage{float}
\usepackage{wrapfig}
\usepackage{placeins}
\usepackage{dblfloatfix}
\usepackage{tabu}
\usepackage{xcolor}
\usepackage{tabularx}
\usepackage{arydshln}
\usepackage[normalem]{ulem}

\def\BibTeX{{\rm B\kern-.05em{\sc i\kern-.025em b}\kern-.08em
    T\kern-.1667em\lower.7ex\hbox{E}\kern-.125emX}}

\usepackage{blindtext}

\let\oldtwocolumn\twocolumn
\renewcommand\twocolumn[1][]{%
    \oldtwocolumn[{#1}{
    \begin{center}
           \includegraphics[width=0.95\textwidth]{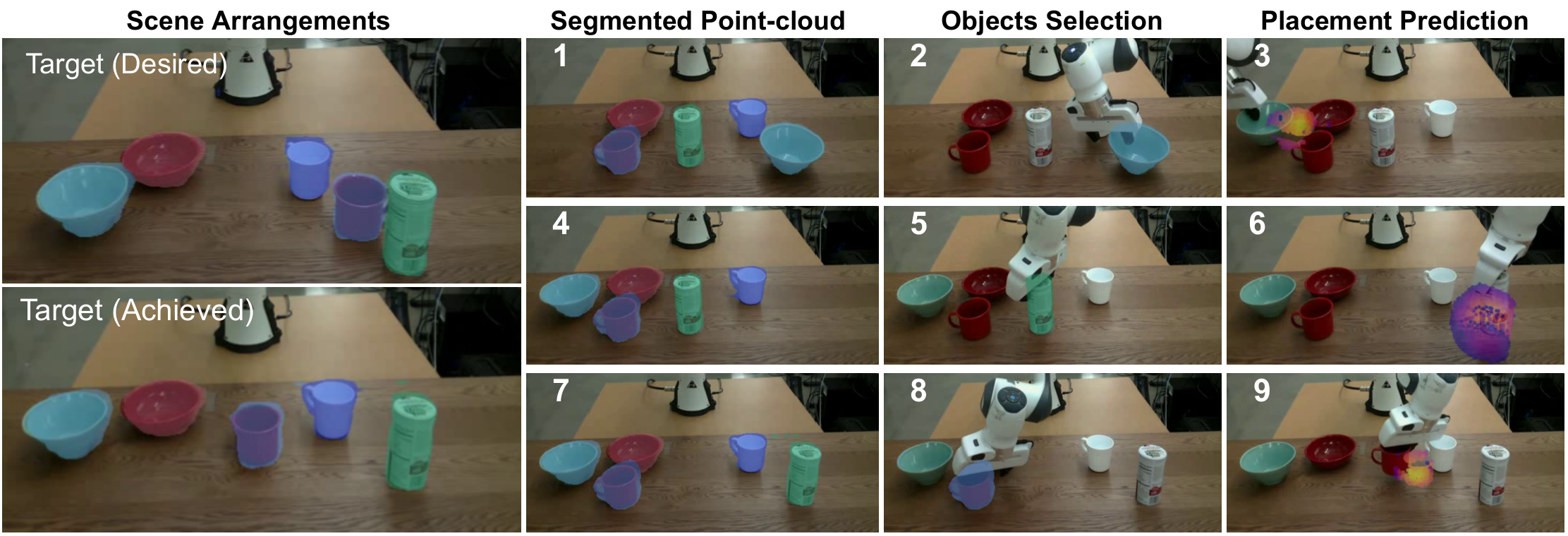}
           \captionof{figure}{
           Neural Rearrangement Planning (NeRP) finds a sequence of pick and place operations to rearrange unknown objects in order to put them in a target arrangement (shown in the top left), choosing both which object to manipulate and where to place it in order to resolve conflicts.
           }
           \label{fig:cover}
        \end{center}
    }]
}

\begin{document}
\title{NeRP: Neural Rearrangement Planning\\ for Unknown Objects}

\author{\authorblockN{Ahmed H. Qureshi$^{1,2}$, Arsalan Mousavian$^1$, Chris Paxton$^1$, Michael C. Yip$^2$, and Dieter Fox$^{1,3}$  }
\authorblockA{$^1$NVIDIA\hspace{15pt}$^2$University of California San Diego\hspace{15pt}$^3$University of Washington\\
a1qureshi@ucsd.edu, amousavian@nvidia.com, cpaxton@nvidia.com, yip@ucsd.edu, dieterf@nvidia.com}}


%

\maketitle

\begin{abstract}
Robots will be expected to manipulate a wide variety of objects in complex and arbitrary ways as they become more widely used in human environments. As such, the rearrangement of objects has been noted to be an important benchmark for AI capabilities in recent years.
We propose NeRP (Neural Rearrangement Planning), a deep learning based approach for multi-step neural object rearrangement planning which works with never-before-seen objects, that is trained on simulation data, and generalizes to the real world. We compare NeRP to several naive and model-based baselines, demonstrating that our approach is measurably better and can efficiently arrange unseen objects in fewer steps and with less planning time. Finally, we demonstrate it on several challenging rearrangement problems in the real world\footnote{Please refer to our supplementary video: \url{https://youtu.be/CJb1IzH94eo}}. 
\end{abstract}
\IEEEpeerreviewmaketitle

\section{Introduction}

Many real-world robotic tasks boil down to pick-and-place, but with a much wider diversity of objects and scenarios than we typically see in the lab. Robots many also encounter cluttered scenes and blocked goals, cases where typically we might need to perform much longer horizon planning to clear objects and move things out of the way.

Rearrangement of unknown objects has been recently identified as a major challenge problem for embodied AI, especially robotics~\cite{batra2020rearrangement}.
While many solutions exist that can solve these problems, particularly throughout the sub-field of task-and-motion planning (TAMP)~\cite{garrett2020integrated}, these often come with a wide range of significant limitations that restrict their utility in the real world.
In particular, these approaches often rely on having known object models.

However, recent advances in deep learning have provided ways for us to weaken these assumptions. Work on deep learning for grasping gives us the ability to grasp unknown objects~\cite{mahler2017dex,mousavian20196}, including in cluttered scenes~\cite{murali20206,martin2021}. A growing body of work also looks at using learned representations for robotic task or motion planning~\cite{srinivas2018universal,hafner2019learning,paxton2019visual,ichter2020broadly}. Others have learned samplers for integration into traditional motion planners, e.g.~\cite{qureshi2020motion, qureshi2020neural}. We can also more accurately segment unknown objects from the world, giving us ways to identify and pick up objects that we have never seen before~\cite{xiang2020learning,xie2020best}. Recently, graph neural networks have been used to represent 3D scenes~\cite{bear2020learning}.


One question that remains, though, is how we can plan to execute sequences of actions including grasps and placements, in order to perform long-horizon rearrangement tasks in unknown environments. Multiple objects might block one another or prevent things from being moved into new positions. In particular, we need to make decisions about which objects to grasp or move when there are multiple options that are viable at any given time. 

In this work, we describe {\bf Ne}ural {\bf R}earrangement {\bf P}lanning (NeRP), an approach for rearranging unknown objects from perceptual data in the real world, as shown in Fig.~\ref{fig:cover}. NeRP represents a scene as a graph over segmented objects. Given a current and the goal scene and object segmentation, it finds an alignment between the two sets of unknown objects in the goal and current image using pre-trained ResNet50~\cite{He2015} features, and uses them to output multi-step rearrangement actions. NeRP is trained using the synthetically generated data in simulation for random object rearrangement.

It uses learned neural networks for choosing the object that needs to be picked and the distribution of possible placements in terms of relative transforms from the current position of the object in the point-cloud space. These operations can be sequenced over time via a sampling-based planning algorithm, which allows us to choose sequences of manipulations in order to perform  rearrangement tasks with unknown objects. At execution time, we use model-free grasping~\cite{martin2021} to pick and use MPPI controller~\cite{danielczuk2020object} to place objects and thereby achieve the specified goal. After each placement action, NeRP observes the scene and re-plans the best course of actions to account for inaccuracies in execution. Fig~\ref{fig:cover} shows execution of object rearrangement on unseen objects with real robot using NeRP.

To our knowledge, this is the first system for end-to-end rearrangement planning for unknown objects. Specifically our contributions are:
\begin{itemize}
    \item a graph neural network approach for computing which objects to move and estimating where they can be placed to complete a rearrangement task,
    \item an algorithm for planning and execution that will reactively select where to place these unknown objects in order to complete the rearrangement task.
\end{itemize}
In addition, we show results both on a simulated object rearrangement task and via real-world robot execution.

\section{Related Work}

Perhaps the most relevant area of work to rearrangement planning is the field of task and motion planning (TAMP).
TAMP is a broad area of study that looks at integrating discrete high-level planning (which objects to grab, which actions to execute) with continuous low level planning (which positions in which to place objects)~\cite{garrett2020integrated}. This area generally relies on model-based methods, using known objects and fully-observable domains, although work has been done to weaken these assumptions, particularly in recent years.
For example, recent work has looked at task and motion planning with partial observability~\cite{garrett2020online}, enabling re-arrangement even if certain objects are not fully visible, and others have used learning to guide task and motion planning~\cite{kim2019learning}. 

Rearrangement planning is a common subset of task and motion planning~\cite{cosgun2011push,king2017unobservable,nam2019planning,labbe2020monte,garrett2020integrated}. It has recently been identified as an interesting challenge area for robotics research~\cite{batra2020rearrangement}. It is interesting in part because it involves long-term planning; recent methods often use Monte Carlo Tree Search or similar methods to explore multiple future possibilities, e.g.~\cite{king2017unobservable,labbe2020monte}. Partial observability is a serious issue, as in many cases objects are partly or wholly occluded~\cite{nam2019planning}, which necessitates special considerations~\cite{garrett2020online}. 

One possible route to building on these is via Motion Planning Networks (MPNet)~\cite{qureshi2020motion, qureshi2020neural, qureshi2020constrained}, which predict a set of waypoints that you can use for classical planning based on sensor data.
Neural Task Graphs~\cite{huang2019neural} propose a way to perform multi-step planning with known objects, but require a demonstration of the correct action sequence.
Another approach is via Deep Affordance Foresight~\cite{xu2020deep}, which learns predictive affordance models to allow completion of longer horizon tasks.

Visual Robot Task Planning~\cite{paxton2019visual} learns an autoencoder style representation for multi-step task planning, but does not generalize to different goals.
Similarly, PlaNET works via deep visual predictions, showing what the possible consequences are for near-horizon actions~\cite{hafner2019learning}. Universal Planning Networks use a learned latent space for motion planning together with a simple gradient descent planner~\cite{srinivas2018universal}.
Other work like Q-MDP net for planning under partial observability~\cite{karkus2017qmdp}, focusing on navigation instead of placement and rearrangement.

Broadly Exploring Local Policy Trees break up task planning into many different learned high level tasks~\cite{ichter2020broadly}, and use an RRT-like approach to navigate between these via a latent space. However, this approach is not applied to the real world in the same way as our approach is, and as it uses a learned latent space it is most likely less general.
Simeonov et al.~\cite{simeonov2020long} propose a method that might be most similar to ours, which looks at a whole sequence of manipulation skills to do task and motion planing with unseen rigid objects. However, they assume a high level task skeleton is given. To the best of our knowledge, NeRP is the first approach that solves long-horizon rearrangement planning tasks with unknown objects. Additional possible related work can be Transporter Networks \cite{zeng2020transporter} which, however, solves significantly simpler tasks requiring one-step predictions and multiple demonstration of each task. Instead, we do long-horizon multi-step predictions with unknown objects from a single target image without any new training.



\section{Problem Definition}
Let $X=\{x_1,x_2, \cdots, x_n\} \in \mathcal{X}$ denote an unordered set of $n$ points each of dimension $d$. Let $\mathcal{M} \in \mathbb{R}^n$ be a point-wise selection operator, i.e., $\mathcal{M}  \times \mathcal{X} \mapsto \mathcal{X}'$, over a given set, where $\mathcal{X}' \subset \mathcal{X}$, and $\Delta \in \mathbb{R}^d$ be a set action operator such that $\mathcal{X} \times \Delta \mapsto \mathcal{X}$.
Therefore, a set-based planner can be defined as a new class of function $\Pi: \mathcal{X} \mapsto \mathcal{M}  \times \Delta$ that determines a sequence of set operators in $ \mathcal{M} \times \Delta$ space to point-wise transform a given set to another desired set. In this paper, we propose NeRP, a set-based planner that outputs a sequence of actions $\{( M_0,\delta_0),( M_1,\delta_1), \cdots, ( M_H,\delta_H)\} \in  \mathcal{M} \times \Delta$ for the initial $X(t)$ and target $X(T)$ unordered sets such that $X(t) \times  \boldsymbol{M} \times \boldsymbol{\delta} \mapsto X(T)$, where $\boldsymbol{M} \in \mathcal{M}$ and $\boldsymbol{\delta} \in \Delta$. We demonstrate the application of our method to rearrangement planning problems for robotics.

Let $\mathcal{Q}$ and $\mathcal{A}$ be the robot configuration and action spaces, respectively. A low-level agent can be defined as a function $\pi^L: \mathcal{Q} \mapsto \mathcal{A}$  that achieves the given robot configurations $ q \in \mathcal{Q}$ by executing a sequence of actions $\boldsymbol{a}_{\{m\}}= \{a_1,a_2, \cdots, a_m\} \in \mathcal{A}$. In our case, we consider the general robot interaction setting, where for a given current $X(t)$ and target $X(T)$ sets, the high-level agent, $\pi^{H} \in \Pi$, at time $t$,  outputs a set action ${(M(t),\delta(t))} \in \mathcal{M} \times \Delta$, leading to an achievable sub-goal set $q_\delta \subset \mathcal{Q}$ for the low-level agent. The low-level agent $\pi^L$ 
executes a sequence of actions $\boldsymbol{a} \in \mathcal{A}$ to achieve $( M(t),\delta(t))$ in the environment and return the control to task planner where it gets new set of observation $X(t+1)$ and replans accordingly. 

\section{Neural Rearrangement Planning}
\begin{figure*}
    \centering
       \includegraphics[width=\textwidth]{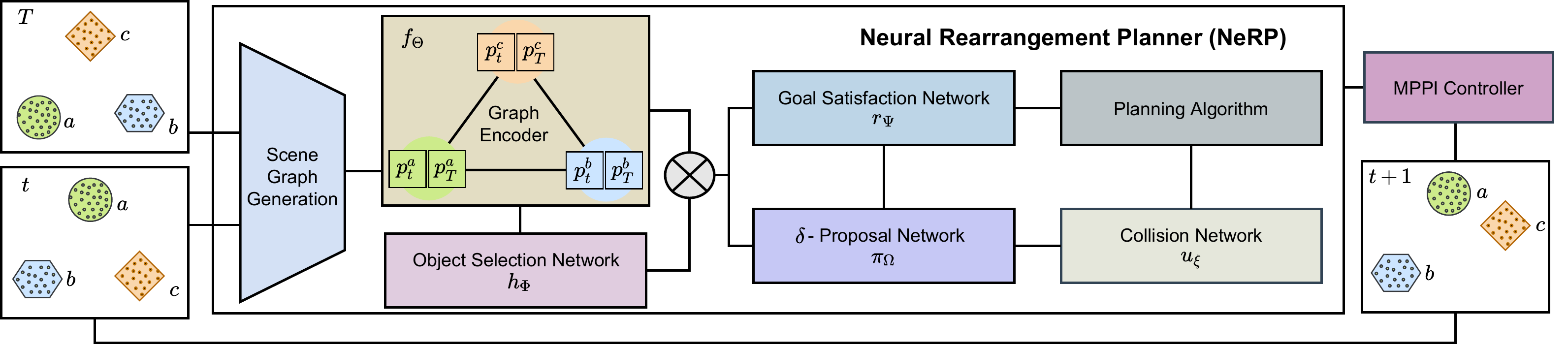}
       \caption{Model architecture overview. We use UCN~\cite{xiang2020learning} to segment out unique objects in the scene, and then compute latent embeddings $\boldsymbol{w}$ for each object alignment in the current and target observation for scene graph generation. Our graph encoder network $f_\Theta$ computes the graph embeddings. Then, at each planning step, we use the object selector $h_\Phi$ to choose which object to move, and the $\delta$-proposal network $\pi_\Omega$ to generate candidate motions. The goal satisfaction network $r_\Psi$ predicts whether or not individual configurations will satisfy the objective, and the collision detection network rejects particular invalid proposals.}
\label{fig:overview}
\end{figure*}   


This section formally presents our approach for Neural Rearrangement Planning (NeRP), which comprises four main components: the object alignment network, the object selection and objects placement prediction networks, and the collision detection network. Fig.~\ref{fig:overview} shows an overview of this model architecture.
We use these networks to perform multi-step planning in order to perform object rearrangement in unknown scenes.

\begin{algorithm}[ht]
\DontPrintSemicolon
$K \gets \mathrm{InstanceSegmentation}(X(t),X(T))$\;
\If{$\boldsymbol{\mathrm{at}} \:t=0$}
{
$h \gets 2 \times |K| \:\textbf{or}\: |K|+1 \hfill //\mathrm{set \:planning \: horizon}$\;
}
\ElseIf{$h==0$}
{
$\boldsymbol{\mathrm{report}} \:\mathrm{out \text{-}of \text{-} planning \text{-} budget}$\;
}

$\boldsymbol{s} \gets \mathrm{ObjsAlignment}(X(t),X(T),K)$\;
$G=(V,E) \gets \mathrm{GraphGen}(X(t),X(T),K,\boldsymbol{s})$\;
$\overline{X}(t),\overline{G} \gets X(t), G \hfill //\mathrm{ make \: a \: backup \: copy}$\; 
$rollouts \gets []$\;
\For{$1$ \KwTo $\mathrm{n\_rollouts}$}
{
$rollout \gets []$\;
$X(t) \gets \overline{X}(t)$\;
$G \gets \overline{G}$\;
\While{$h>0$}
{
$\boldsymbol{z}_{\{N\}} \gets f_\Theta(G) \hfill //\mathrm{graph \: encodings}$\;
$\rho_{\{N\}} \gets h_\Phi(\boldsymbol{z}_{\{N\}})$\;
$i \sim \mathrm{Mult}(\rho_{\{N\}}) \hfill //\mathrm{graph \: node \: sampling} $\;
$\delta_{\{B\}} \gets \pi_\Omega(\boldsymbol{z}^i_{\{B\}}) \hfill //\mathrm{set \: action \: generation}$\;
$\hat{X}^i_{\{B\}}(t)=X^i_{\{B\}}(t)+\delta_{\{B\}}$\;
$N'=N\backslash i \hfill //\mathrm{unselected \: node \: indices}$\;
$\hat{X}^{N'}_{\{B\}}(t)=X^{N'}_{\{B\}}(t)$\;
$\boldsymbol{x}_{\{N' \times B\}} \sim \{(\hat{X}^i_{\{B\}}, \boldsymbol{0}) \cup (\hat{X}^j_{\{B\}}, \boldsymbol{1})\} \: \: \forall j \in N'$\;
$ids \gets u_\xi(\boldsymbol{x}_{\{N' \times B\}})>\epsilon$\;
$\delta_{\{B'\}}=\delta_{\{B\}}[ids] \hfill //\mathrm{collision\text{-}free \: actions}$\;
$v_{\{B'\}}\gets r_\Psi(\boldsymbol{z}^i_{\{B'\}},\delta_{\{B'\}}) \hfill //\mathrm{set\text{-}action \: scores}$\;
$j \gets \mathrm{argmax} \ v_{\{B'\}} \hfill //\mathrm{best \: action}$\;
$V^i(t) \gets V^i (t)+ \delta^j \hfill //\mathrm{update\: graph\: vertex}$\; 
$X^i(t) \gets X^i(t)+\delta^j \hfill //\mathrm{update\: point\text{-}cloud}$\; 
$e\gets \mathrm{ComputeError}(V)$\;
$rollout \cup (i,\delta_{\{B'\}},j,e) $\;
}
$rollouts \cup rollout$\;
}
$(i,\delta_{\{B'\}},j)\gets \mathrm{BestRollout}(rollouts)$\;
$h \gets h-1$\;
$X^i_{best} \gets X^i(t)+\delta_j \hfill //\mathrm{best \: placement}$\;
$\boldsymbol{p}_{best}\gets \mathrm{ComputeMean}(X^i_{best})$\;
$X^i_{all} \gets X^i_{\{B'\}}(t)+\delta_{\{B'\}} \hfill //\mathrm{all \: placement \: points} $\;
$X^i_{\mathcal{B}}\gets k\text{-}\mathrm{Neighbors} (\boldsymbol{p}_{best},X^i_{all})$\;
$ c_{map} \gets \mathrm{ComputeCost}\big(\boldsymbol{p}_{best} ,X^i_{\mathcal{B}}\big)$\;
\Return $X^i_{\mathcal{B}}, c_{map}$
\caption{NeRP $\big(X(t)$, $X(T)\big)$}
\label{nerp}
\end{algorithm}

\subsection{Neural Networks} 

\textbf{Objects Alignment and Graph Generation:}
The objects alignment module determines the individual objects' correspondence in the current $X(t)$ and target $X(T)$ scene point-clouds for the graph generation. We use Unknown Objects Instance Segmentation (UCN)~\cite{xiang2020learning}
to 
extract specific object's RGB-information from the given scene and it's corresponding point clouds $X^i \subset X$, where $i$ denotes the selected object's index. The RGB information for each individual objects are passed through pre-trained Resnet model~\cite{He2015} to extract their latent features $\boldsymbol{w} \in W$ for computing L2 norm based objects  similarity scores $\boldsymbol{s}$, an $N \times N$ matrix with $N \in \mathbb{N}$ denoting the number of instance segmentations in the current and target scenes.

The graph generator takes the current and target scene point clouds with their segmentation labels and the similarity matrix $\boldsymbol{s}$ as an input and generates an undirected graph  $G=(V,E) \in \mathcal{G}$. Each vertex $V^i \in V$ of the graph is computed as follows. Objects in the current observation and target observation are assigned to each other by minimizing the assignment cost using Hungarian method. For instance, in Fig \ref{fig:overview}, object $a_t$ from the current scenario (at time $t$) is paired with object $a_T$ from the target scenario (at time $T$) using their feature-based similarity scores $s(a_t,a_T)$. Once object assignment is computed, the center of observed object's point cloud $p_t \in  \mathbb{R}^3$ and $p_T \in \mathbb{R}^3$ are concatenated to represent $V^i \in \mathbb{R}^6$, the features for a graph vertex.

  
\textbf{Graph Encoder Network:}
Our graph encoder network is based on a Higher-order Graph Neural Networks \cite{morris2019weisfeiler}, known as $k$-GNNs, that takes the graph $G=(V,E)$ and hierarchically learns the objects' latent embeddings, $\{\boldsymbol{z}^i; \: \: \forall i \in [0,N]\}$. We use local $k$-GNNs, more precisely 1-2-3-GNNs with max-pooling aggregation operator, where each graph layer performs message-passing between individual vertices and local subgraph structures, denoted as $g \subset G$, along the hierarchy from $k=1$ to $k=3$, i.e.,
\begin{equation}
f_k^t(g)=\sigma\bigg(f_k^{t-1}(g) \cdot \Theta^t_1 +\max_{g\in \mathcal{N}_L(s)}\big(f_k^{t-1}(g)\cdot \Theta^t_2\big)\bigg)
\end{equation}
where $N_L$  denotes the local-neighborhood of subgraph $g$ (for more details, refer to \cite{morris2019weisfeiler}), $\Theta^t_1$ and $\Theta^t_2$ are the learnable weight parameters, and $\sigma$ is the component-wise non-linear function ReLU. In our problem setting, $k$-GNNs with max-pooling operators outperformed vanilla GNNs \cite{wu2020comprehensive}, as the former captures the fine to coarse level structure of the given current-target aligned scene graph which is crucial to have scene-aware node embeddings. Note that our object alignment network captures across the scene correspondence between objects, whereas our graph encoder network captures the overall current and target placement structure (Fig. \ref{fig:overview}).

\textbf{Object Selection Network:}
Our Object Selection Network, $h_{\boldsymbol{\Phi}}: \mathcal{G} \mapsto \mathcal{M}$, is an object-centric neural model, with parameters $\boldsymbol{\Phi}$, that takes the individual graph node embeddings $\boldsymbol{z}_{\{N\}}$, representing current-target scene graph, and predicts their selection scores $\rho_{\{N\}}$, where $\rho^i\in [0,1]$, i.e.,
\begin{equation}
\rho_{\{N\}} \gets h(\boldsymbol{z}_{\{N\}}; \boldsymbol{\Phi}) 
\end{equation}
We convert all selection scores $\rho_{\{N\}}$ into probabilities to parameterize a multinomial distribution. A graph node index $i$ is sampled from this distribution during planning which denotes a particular object-pairs embedding in the scene graph. Note that the index $i$ also maps to the instance segmentation label, thus resulting into a subset selection operator $M_i \in \mathcal{M}$. The selected object pair's graph node embedding $\boldsymbol{z}^i$ is then used to predict the next relative transformation for the subset $M_i \times X(t) \mapsto X^i(t)$. Furthermore, we train this module using the binary-cross entropy loss as:
\begin{equation}
l_{\Phi,\Theta}=-\cfrac{1}{N}\sum_i y^i \cdot \log(h_{\mathrm{\Phi}}(\boldsymbol{z}^i)+(1-y^i)\cdot \log(1-h_{\mathrm{\Phi}}(\boldsymbol{z}^i)) 
\end{equation}  
where $y^i$ is the binary label from demonstrations indicating a graph node to be selected. 
 
\textbf{$\delta$-Proposal Network:}
The $\delta$-Proposal Network $\pi_\Omega: \mathcal{X} \mapsto \Delta$, with parameters $\Omega$, is a stochastic neural model that takes the selected node embeddings $\boldsymbol{z}^i_{\{B\}}$ and outputs a variety of candidate translations $\delta^i_{\{B\}} \in \Delta$ that would move the object to a potential new placement region, i.e., 
\begin{equation}
\delta^i_{\{B\}} \gets \pi(\boldsymbol{z}^i_{\{B\}}; \Omega)
\end{equation}
where $\delta^i_{\{B\}}$ contains a variety of $\delta$ actions of size $B$ and $\boldsymbol{z}^i_{\{B\}}$ contains $B$ replicas of $\boldsymbol{z}^i$. These $\delta^i_{\{B\}}$ actions are applied to $B$ replicas of selected object's current point-cloud $X^i_{\{B\}}(t)$ which leads to its next placement in the point-cloud space, i.e.,

\begin{equation}
X^i_{\{B\}}(t+1)=
\begin{bmatrix}
X^i_{1}(t) \\
\vdots\\
X^i_{B}(t) 
\end{bmatrix} +
\begin{bmatrix}
\delta^i_1 \\
\vdots\\
\delta^i_B 
\end{bmatrix}
\end{equation}
Furthermore, this model obtains its stochasticity from Dropout layers \cite{srivastava2014dropout} applied to the networks' linear layers with probability $p \in [0,1]$ during execution. For the given Graph $G=(V,E)$ and the true next step delta $\overline{\delta}$ from demonstrations, this module is trained by minimizing the following: 
\begin{equation}\label{mse1}
l_{\Omega,\Theta}(M,f_\Theta(G),\overline{\delta})=\big\|\rho_\Omega \big(M \circ f_\Theta(G)\big)-\overline{\delta}\big\|_2
\end{equation}
where $M$ is a mask operator which selects the graph node embedding that corresponds to the given true delta $\overline{\delta}$ label only. 

\textbf{Goal Satisfaction Network:}
Our goal satisfaction network is a value function that scores the given actions for their ability to accomplish the given target arrangements. It is a neural function $r_\Psi: \mathcal{G} \times \Delta \mapsto [0,1] $, with parameters $\Psi$, that takes a selected graph node embedding and the given action $\delta$ as an input and outputs goal satisfaction scores, i.e.,
\begin{equation}
\hat{y}_r\gets \sigma\big(r_\Psi(M \circ f_\Theta(G),\delta)\big)
\end{equation} 
where $\hat{y}_r \in [0,1]$, $M$ is a mask operator that selects a graph node embedding corresponding to the given action $\delta$, and  $\sigma$ is a sigmoid function to squash predicted scores to $[0,1]$. The $\hat{y}_r = 1$ indicates that the given action $\delta$ can take the selected object to its target location, and $\hat{y}_r = 0$ indicates otherwise. Furthermore, this function is optimized through binary-cross entropy loss as follows:
\begin{equation}
l_{\Psi, \Theta}(M,f_\Theta(G),y_{r})=- y_r \cdot \log(\hat{y}_r)-(1-y_r)\cdot \log(1-\hat{y}) 
\end{equation} 
where $y_r$ denotes the true label.

\textbf{Collision Network:}
Our collision detection network $u_\xi: \mathcal{X} \mapsto [0,1]$, with parameters $\xi$, uses a number of PointNet++ set-abstraction layers \cite{qi2017pointnet++} to detect the intersection between any two given sets ${X_0,X_1} \subset X$. In our setting, $X_0$ and $X_1$ can be any arbitrarily selected object point-clouds with their feature labeled as $Y_0=\boldsymbol{0}$ and $Y_1=\boldsymbol{1}$, respectively. Our collision network's input is a subset sampled from a joint-set of given point-clouds and their feature masks, i.e., $\boldsymbol{x} \subset (X_0,Y_0) \cup (X_1,Y_1)$. The feature mask indicates element-wise point-cloud correspondence to the given sets. The network predicts scores $\rho_c \in [0,1]$ indicating degeree of sets' intersections. We train our collision-checker independently from other NeRP models using the BCE loss with training samples containing both intersecting and disjoint point-cloud sets.
\begin{figure}[ht]
\includegraphics[width=0.49\textwidth]{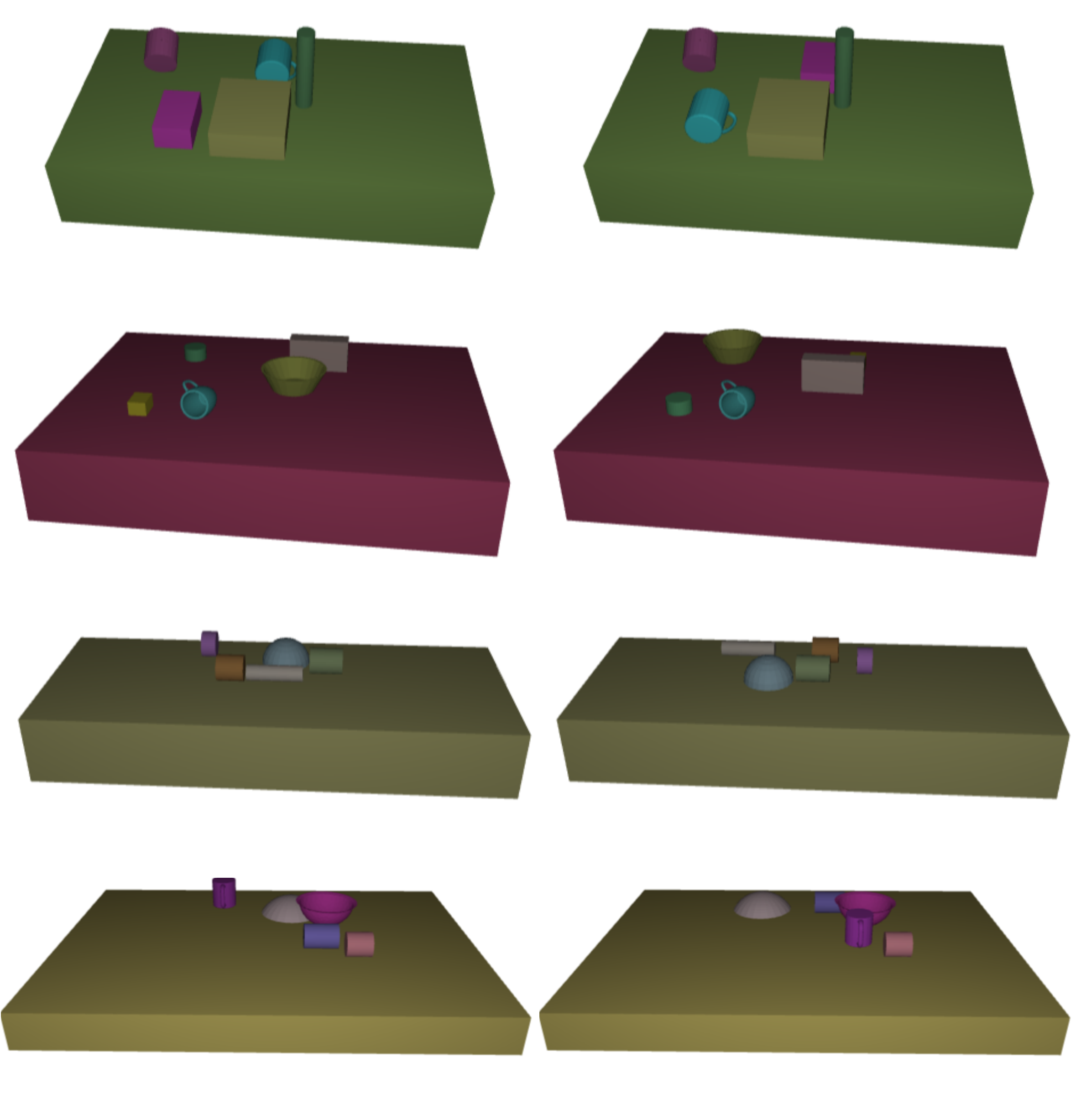}
\caption{Examples of generated data. Objects are randomly placed on the table, and we chose different random motions as well.}
\label{fig:datagen}
\end{figure}
\subsection{NeRP Training}
We train our core models, i.e., graph encoder $f_\Theta (\cdot)$, node selector $h_\Phi(\cdot)$, $\delta$-proposal network $\pi_\Omega (\cdot)$, and goal satisfaction evaluator $r_\Psi(\cdot)$, jointly in an end-to-end manner by optimizing the following objective:
\begin{equation}\label{mse2}
\cfrac{1}{N_\mathcal{M}} \sum_{\tau \sim \mathcal{M}}  l_{\Phi,\Theta}(\boldsymbol{z}_{\{N\}},y_{\{N\}})+l_{\Omega,\Theta}(\boldsymbol{z}^i,\overline{\delta}^i)+l_{\Psi, \Theta}(\boldsymbol{z}^i,\overline{\delta}^i,y_{r}),
\end{equation}
where $\tau=(G,y_{\{N\}},\overline{\delta}^i, M, y_r)$ denotes a training sample from demonstration data $\mathcal{M}$, comprising a current-target paired scene graph $G$ with its node embeddings $\boldsymbol{z}_{\{N\}}=f_\Theta(G)$. It also includes the node selection labels as $y_{\{N\}}$, and a differentiable mask $M$ operator to select a desired node embedding $\boldsymbol{z}^i=M\circ\boldsymbol{z}_{\{N\}}$. Furthermore, we also provide a desired next action $\overline{\delta}^i$ and its goal satisfaction value $y_r$ for the selected node. The graph encoder is learned end-to-end with other core models to capture scene embedding useful for the overall rearrangement planning process whereas the collision network is trained independently.

\subsection{NeRP Planning Algorithm}
Algorithm~\ref{nerp} describe our multi-step planning algorithm 
for efficiently solving tabletop rearrangement tasks. 
For the given observations, $X(t)$ and $X(T)$, our method begins by computing their instance segmentation $K$ using UCN \cite{xiang2020learning}. A current-target aligned scene graph $G$ is then instantiated using those segmented observations and the objects similarity scores $\boldsymbol{s}$ (Lines 6-7).

Given an observation graph $G=(V,E)$, our multi-step predictive planning approach imagines $n$ action sequences out to horizon $h$, stores them into a buffer $rollouts$, and executes the first action through a low-level controller of the best-selected planning sequence. In simulated environments, we set horizon length $h$ to $2\times |K|$, whereas in real-world experiments, we further limit the planning budget to $|K|+1$. Furthermore, we use a Model Predictive Path Integral (MPPI) controller based on learned collision avoidance models~\cite{danielczuk2020object} to perform the given placement actions leading to the next observation graph for the planning.
After each pick-and-place action is executed, the horizon $h$ is decremented and we replan.
If controller fails to execute an action (e.g. it fails to grasp or drops the object early) the plan horizon is incremented again to give another opportunity to the planner for re-planning.

In a planning sequence, the graph encoder, $f_\Theta$, outputs the graph nodes' embeddings $\boldsymbol{z}_{\{N\}}$. The function $h_\Phi$ takes graph embeddings and predicts $\rho_{\{N\}}$ which parametrize a Multinomial distribution for sampling an index $i \in [0, N]$. For a selected graph node's embedding $\boldsymbol{z}^i$, multiple replicas are fed to our stochastic $\delta$-proposal network, $\pi_\Omega$, to determine the various next step actions $ \delta_{\{B\}}$ for the placements of the object $i$ (Line 15-18).

Each of the candidate placements of the selected object in $\hat{X}^i_{\{B\}}(t)$ forms a new scenario. To determine the best next-step scene arrangement, we translate the object point-cloud according to the sampled $\delta$ and use our collision and goal-satisfaction networks. The collision-network, $u_\xi$, takes all possible next step objects point-clouds and returns action indices, $ids$, leading to all collision-free next-step arrangements (Line 22-24). The goal satisfaction network takes these collision-free placement actions, $\delta_{\{B'\}}$, and provides the scores, $v_{\{B'\}}$, to select the best move for simulating the next-step using the graph $G$ and current point-cloud $X(t+1)$ (Line 25-28).

Furthermore, for each planning step, we compute a L2 norms based error $e$ between the updated graph vertices $V(t)$ and $V(T)$, indicating difference between current simulated scene and the final arrangement. All this planning step information, including error $e$, is stored in the rollout buffer. Once all sequences are unrolled, the first step of the best-unrolled sequence, based on minimum error $e$, is selected to determine best object $i$ and the placement points, $X^i_{all}$, and a placement cost map $c_{map}$ for the object $i$. To select $X^i_{all}$ and compute their cost-map $c_{map}$, we calculate the centroid, $p_{best}$, of the next best placement point-cloud $X^i_{best}$, and select all points in $X^i_{all}$ that are within a ball of radius of $p_{best}$.

The cost values are calculated based on the distance between $p_{best}$ and all placement locations $X^i_\mathcal{B}$ (Line 38). These placement locations with cost map are given to the MPPI low-level controller for robot execution. The controller generates grasps for object $i$  using~\cite{martin2021} and executes the motions using~\cite{danielczuk2020object}. Once the object is lifted, it chooses the placement with minimum $c_{map}$ for which there is a collision-free, kinematically feasible path.

\begin{figure*}[ht]
    \centering
       \includegraphics[width=\textwidth]{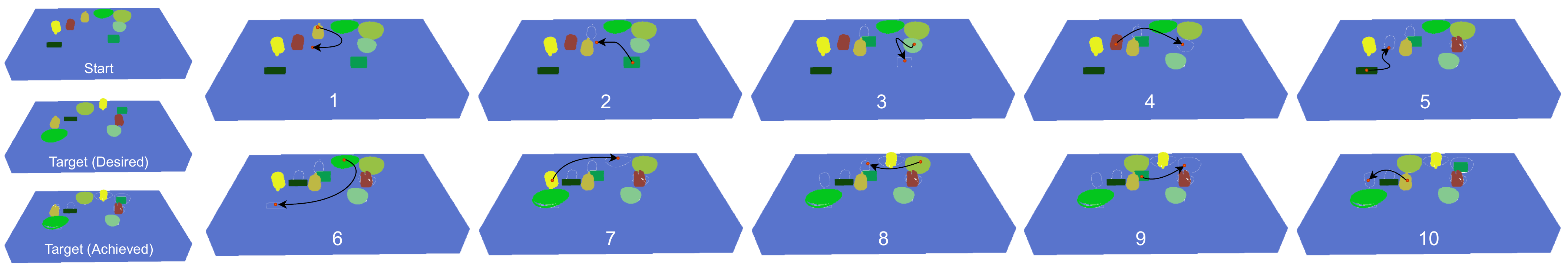}
       \caption{An example plan rollout showing how NeRP chose to move objects around in order to get between two goal states with very different arrangements of obstacles. In this case, it took 10 steps to get to the goal state.}
\label{fig:syn1}
\end{figure*}

\subsection{Data Generation}
\label{sec:experiments-data}
To train the NeRP models, we generate random synthetic scene data rendered with variable camera poses. All scenes contained randomly selected five objects, initially scattered all over the tabletop of changing dimensions (as shown in Fig. \ref{fig:datagen}).
Each scenario's target arrangement was determined by randomly swapping the objects' placements in the initial scene, so that each object's goal location is blocked by a random other object.
Hence, transitioning from an initial to target scene requires some items to be moved to another empty location, which we call \emph{storage}, to vacate the occupied position.
We generate the intermediate placement action using a model-based expert rearrangement planner (as described in Section V). The expert planner generates intermediate scene sequences from which we determine the step-wise relative scene transformations $(M,\overline{\delta}) \in \mathcal{M} \times \Delta $ for training our model-free, set-based rearrangement planner. Following this procedure, we gathered a dataset comprising seven thousand re-arrangement task problems with their intermediate planning sequences. 
 
\subsection{Model Architectures}
This section presents the model-architectures of NeRP. Our networks were implemented using PyTorch~\cite{NEURIPS2019_9015} and PyTorch Geometric Library~\cite{Fey/Lenssen/2019}.  
\begin{itemize}
    \item \textbf{Graph Encoder:} It contains two layers of $k$-GNNs~\cite{morris2019weisfeiler} each with hidden feature size of 512 with max operator.

    \item \textbf{Object Selection Network:} The input graph node features of size 512 are given to two fully connected (FC) layers, followed by ReLU, and an output layer of size 512, 256, and 1, respectively. A sigmoid layer squashes the output, i.e., FC(512, 512) $\rightarrow$ ReLU$\rightarrow$FC(512, 256) $\rightarrow$ ReLU$\rightarrow$FC(256, 1) $\rightarrow$ Sigmoid.

    \item \textbf{$\delta$-proposal Network:} The input node features of size 512 are passed through four FC layers, followed by a ReLU, and an output layer with 512, 384,256,128, and 3 units, respectively. The first three FC layers also contain Dropout with a probability of 0.5, i.e., FC(512, 512) $\rightarrow$ ReLU$\rightarrow$ Dropout (0.5)$\rightarrow$FC(512, 384) $\rightarrow$ ReLU$\rightarrow$ Dropout (0.5)$\rightarrow$ FC(384, 256) $\rightarrow$ ReLU$\rightarrow$ Dropout (0.5)$\rightarrow $FC(256, 128) $\rightarrow$ ReLU$\rightarrow$ FC(128, 3)
    \item \textbf{Goal Satisfaction Network:} This network follows the same structure as the object selection network. 
        
    \item \textbf{Collision Detection Network:} The collision network follows a structure of PointNet++~\cite{qi2017pointnet++} classifier with set abstraction (SA) and FC, followed by ReLU, layers. The SA layer's inputs are defined in the following format (sampling ratio, radius, MLP), where MLP denotes Multi-Layer Perceptron. Followed by the SA layers is a GlobalSA layer, i.e., SA(0.5, 0.4, MLP([4, 32, 32, 64])) $\rightarrow$ SA(0.25, 0.6, MLP([64 + 3, 128, 128])) $\rightarrow$ GlobalSA(MLP([128 + 3, 256, 512])) $\rightarrow$
FC(512, 256) $\rightarrow$ ReLU$\rightarrow$FC(256, 128)$\rightarrow$ReLU$\rightarrow$FC(128, 1)$\rightarrow$Sigmoid.
\end{itemize}

\subsection{Hyperparameters} We use Adagrad optimizer~\cite{JMLR:v12:duchi11a} with learning rate$=0.01$, learning rate decay$=0$, initial accumulator value$=0$, weight decay $=0$, and eps$=1e-10$.

\section{Results}
We performed four sets of experiments. First, we tested our method on unseen synthetic data against various classical baselines. Second, we show generalization of our method on object rearrangement for unseen number of objects. Third, we do ablation study to evaluate effect of each component of NeRP. Fourth, we demonstrate our method's sim-to-real generalization performance on real-world object rearrangement tasks with different sets of unseen objects and unseen rearrangement tasks. We use the following metrics for quantitative comparison of different methods:
\begin{itemize}
    \item {\bf Success Rate} indicates the percentage of successfully solved unseen arrangement problems; an object arrangement is considered successful if the maximum displacement for each object does not exceed $5mm$.
    \item {\bf Planning Steps} measures the number of steps required to rearrange the objects from source configuration to target configuration.
    \item {\bf Final Error} measures the average L2 distance between the desired target arrangement and the actual arrangement achieved by a given planner. 
\end{itemize}
Note that success rate is computed over all the rearrangement scenarios whereas planning steps and final error are computed only on successful rearrangements.

\subsection{Algorithm Comparison}
To validate our approach, we first tested on 500 simulated unseen object rearrangement problems that were generated with 5 objects following the procedure in Section IV-D.  Table~\ref{tab:comparison} shows how our method compares to multiple baseline methods. These baselines include:

\textbf{Model-based expert:} The expert planner is a model-based planning approach that uses objects' unique ids, transformations, table mesh, object meshes, and FCL~\cite{6225337} collision-checker for planning. For this model, we set the same step length limit as NeRP. It randomly selects an object to move that is not at its goal position, and then checks to see if its target location is empty or occupied. If the target location is occupied, it will move the occupying object to free space, chosen by randomly sampling a feasible \emph{storage} location on the table. If the target location is free, it will instead move the object to the goal. This process is repeated until all the objects are at their goal positions.

\textbf{Classical (heuristic, parallelized):}
This is a model-free version of the ``model-based expert'' planner.
It uses instance segmentation and our object alignment model to match objects between the current and target scenes, and follows a following greedy heuristics. Once objects pair are computed, it randomly selects an object, checks if its target location is occupied or empty, and moves the occupying object to storage by sampling multiple placements in parallel and rejecting colliding placement positions using our collision-net $u_\xi$ instead of using FCL. Once the target location is free, it moves the selected object to its target, and proceeds by randomly selecting another object and repeating the process. 

\begin{table}
\centering 
\scalebox{0.75}{
\begin{tabular}{cccc}\toprule
\multirow{2}{*}{Algorithms}&\multicolumn{3}{c}{Performance Metrics}\\ \cmidrule{2-4}
&\multicolumn{1}{c}{Success rate $(\%)$ $\uparrow$}&\multicolumn{1}{c}{Planning steps $\downarrow$}&\multicolumn{1}{c}{Final error $\downarrow$}\\\midrule

\multirow{1}{*}{Model-based expert}& \multirow{1}{*}{$90.67 \pm 0.60$} & \multirow{1}{*}{$8.41 \pm 2.61$}  &\multirow{1}{*}{$\mathbf{0.0 \pm 0.0}$}\\
\multirow{1}{*}{NeRP (Ours)}& \multirow{1}{*}{$\mathbf{94.56 \pm 0.73}$} & \multirow{1}{*}{$\mathbf{7.01 \pm 2.10}$}  &\multirow{1}{*}{$0.019 \pm 0.013$}\\
\multirow{1}{*}{Classical (heuristic, parallelized)}& \multirow{1}{*}{$68.20 \pm 0.79$} & \multirow{1}{*}{$13.60 \pm 5.99$}  &\multirow{1}{*}{$0.023 \pm 0.017$}\\
\multirow{1}{*}{Classical (random, parallelized)}& \multirow{1}{*}{$59.23 \pm 1.32$} & \multirow{1}{*}{$42.80 \pm 60.63$}  &\multirow{1}{*}{$0.019 \pm 0.011$}
\\ \bottomrule
\end{tabular}}
\caption{Comparison between NeRP and several classical baselines. NeRP produces shorter, more accurate plans than baseline methods.} \label{tab:comparison}
\end{table}

\begin{figure*}
    \centering
       \includegraphics[width=0.99\textwidth]{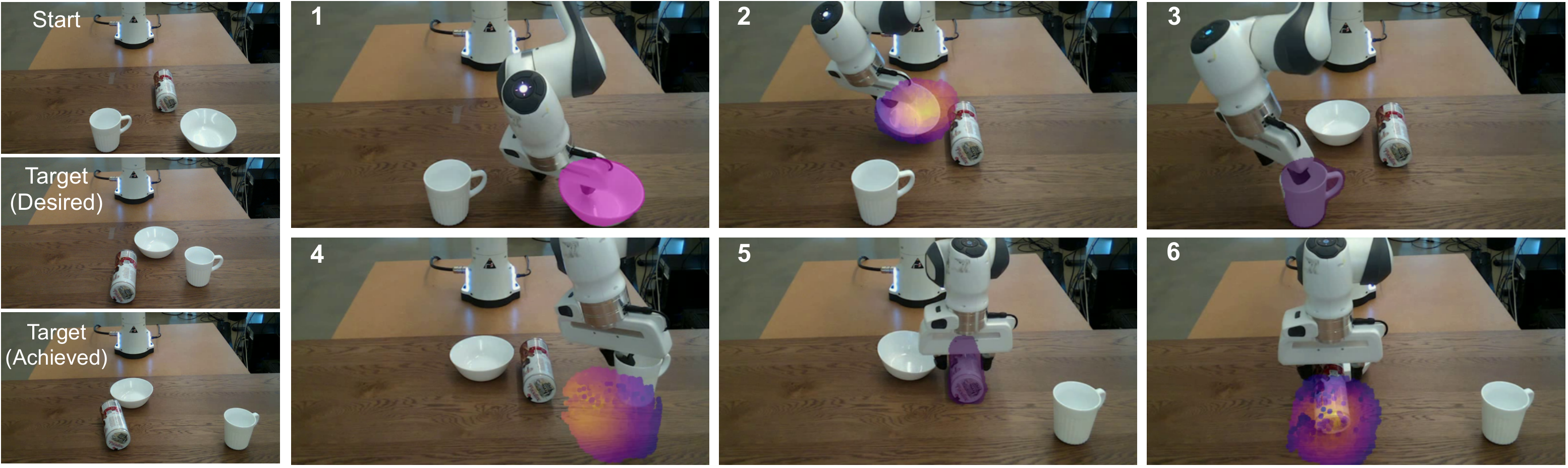}
    \caption{Example of a planning sequence. The robot repeatedly selects which object to move and either moves it to the appropriate goal position or to a \emph{storage} position in order to enable future execution.}
\label{fig:swap1}
\end{figure*}

\textbf{Classical (random, parallelized):}
The random baseline plans along multiple sequences, and executes the first action of the best selected sequence, following the algorithm given by Alg.~\ref{nerp}. The best sequence is selected based on minimum error, $e$, from the target arrangement. However, unlike, NeRP, this baseline randomly chooses an object and performs a sequence execution like a classical (heuristic, parallelized) approach mentioned above.

\noindent \textbf{Comparison to classical methods:} Table \ref{tab:comparison} shows how NeRP out-performs all baseline methods, including the model-based expert that uses all environment state information.
This is for several reasons.
First, the random placements can often be in-collision, especially in cases with a small tabletop area. Second, model-free baselines diverge in cases with noisy feature-based objects' pairing, which is inevitable with learning-based feature matching approaches. Third, in the multi-step, multi-sequence classical planning cases, executing the first action and replanning the sequences again instead of greedily utilizing each action step takes a longer time to converge.

\begin{table}
\centering 
\scalebox{0.85}{
\begin{tabular}{cccc}\toprule
\multirow{2}{*}{Number of  Objects}&\multicolumn{3}{c}{Performance Metrics}\\ \cmidrule{2-4}
&\multicolumn{1}{c}{Success rate $(\%)$ $\uparrow$}&\multicolumn{1}{c}{Planning steps $\downarrow$}&\multicolumn{1}{c}{Final error $\downarrow$}\\\midrule

\multirow{1}{*}{3 Objects}& \multirow{1}{*}{$98.25 \pm 0.57$} & \multirow{1}{*}{$4.58 \pm 0.82$}  &\multirow{1}{*}{$0.039 \pm 0.036$}\\
\multirow{1}{*}{4 Objects}& \multirow{1}{*}{$97.60 \pm 1.20$} & \multirow{1}{*}{$5.70 \pm 1.38$}  &\multirow{1}{*}{$0.027 \pm 0.025$}\\
\multirow{1}{*}{6 Objects}& \multirow{1}{*}{$98.09 \pm 0.40$} & \multirow{1}{*}{$8.69 \pm 2.15$}  &\multirow{1}{*}{$0.013 \pm 0.013$}\\
\multirow{1}{*}{7 Objects}& \multirow{1}{*}{$90.62 \pm 1.03$} & \multirow{1}{*}{$9.47 \pm 2.23$}  &\multirow{1}{*}{$0.011 \pm 0.008$}\\
\multirow{1}{*}{8 Objects}& \multirow{1}{*}{$87.50 \pm 2.50$} & \multirow{1}{*}{$10.72 \pm 2.08$}  &\multirow{1}{*}{$0.010 \pm 0.008$}
\\ \bottomrule
\end{tabular}}
\caption{Generalization of NeRP to different number of objects. The number of planning steps required increases more or less linearly, though the success rate drops slightly as we add more objects. Note that NeRP is trained on random rearrangements of 5 objects.} \label{tab:scaling}
\end{table}

\begin{table}
\centering 
\scalebox{0.85}{
\begin{tabular}{cccc}\toprule
\multirow{2}{*}{Algorithms}&\multicolumn{3}{c}{Performance Metrics}\\ \cmidrule{2-4}
&\multicolumn{1}{c}{Success rate $(\%)$ $\uparrow$}&\multicolumn{1}{c}{Planning steps $\downarrow$}&\multicolumn{1}{c}{Final error $\downarrow$}\\\midrule

\multirow{1}{*}{NeRP}& \multirow{1}{*}{$94.56 \pm 0.73$} & \multirow{1}{*}{$7.01 \pm 2.10$}  &\multirow{1}{*}{$0.019 \pm 0.013$}\\
\multirow{1}{*}{w/o Dropout}& \multirow{1}{*}{$36.87 \pm 0.24$} & \multirow{1}{*}{$8.23 \pm 3.07$}  &\multirow{1}{*}{$0.019 \pm 0.012$}\\
\multirow{1}{*}{w/o OS $h_\Phi$}& \multirow{1}{*}{$30.12 \pm 1.10$} & \multirow{1}{*}{$11.82 \pm 3.30$}  &\multirow{1}{*}{$0.025 \pm 0.021$}\\
\multirow{1}{*}{w/o GS $r_\Psi$}& \multirow{1}{*}{$48.21 \pm 0.68$} & \multirow{1}{*}{$14.84 \pm 1.14$}  &\multirow{1}{*}{$0.016 \pm 0.010$}
\\ \bottomrule
\end{tabular}}
\caption{Analysis of the effects of ablation of various components of the network. Removing stochasticity, the object selection network or the goal selection network has a significant negative effect on performance.} \label{tab:ablation}
\end{table}

\begin{figure*}[t]
    \centering
       \includegraphics[width=\textwidth]{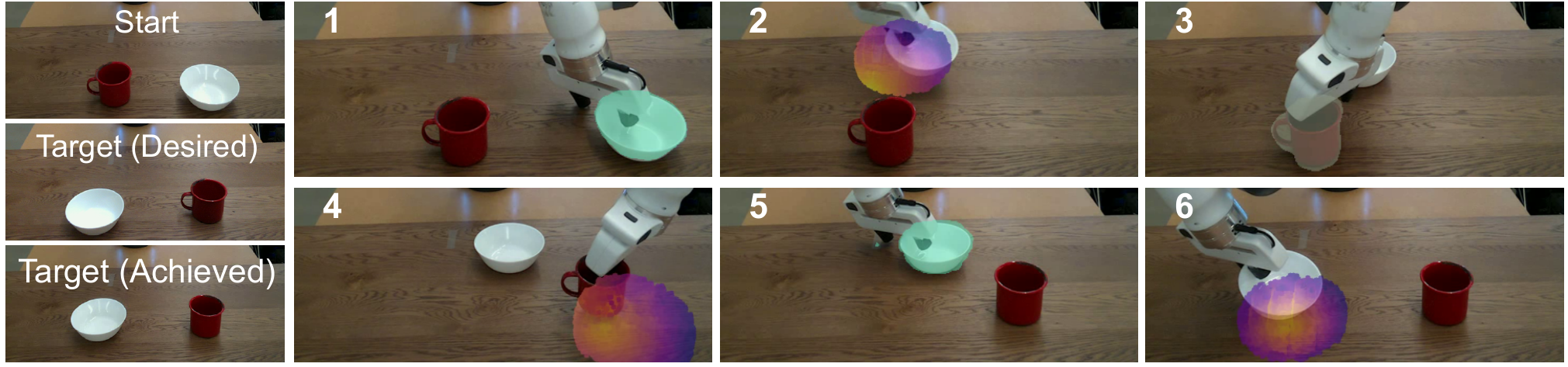}
       \caption{Swapping an unseen mug and bowl using NeRP: For the given $X(\mathrm{Start})$ and $X(\mathrm{End})$ arrangements, NeRP generates an encoded scene graph using which the object selection network ($h_{\Phi}$) selects an object in the given scenarios (e.g., 1, 3 \& 5). The $\delta$-proposal network ($\pi_\Omega$) predicts $\boldsymbol{\delta} \in \Delta$ for the selected object, leading to its next placement region with a cost map $c_{map}$ ( e.g., 2, 4 \& 6). The $c_{map}$ is originated based on the goal satisfaction network's ($r_\Psi$) scores, $\boldsymbol{v}$, to indicate the robot with the best placement locations during execution.
       }
\label{fig:swap2}
\end{figure*} 
\noindent \textbf{Generalization to different numbers of objects:}
Table \ref{tab:scaling} shows NeRP performance over 500 scenarios with a number of objects ranging from two to eight, whereas NeRP training dataset contained scenes with only five objects. Note that we show final errors for only the successful task completions; due to this, the reported errors go down together with the success rates as the number of objects in the scene increases. Fig. \ref{fig:syn1} shows the NeRP execution in a scene arrangement task with eight objects. From these results, we can see that NeRP's object-centric planning is robust to scene clutter and can efficiently sample multiple \emph{storage} locations along the planning sequence for achieving a complex target arrangement setup. 

\subsection{Ablation Studies}

We performed a set of experiments 
ablating various components of our method.
In particular, we look at the importance of the stochastic Dropout added to our $\delta$-proposal network $\pi_\Omega$, the object selection network $h_\Phi$, and the goal satisfaction network $r_\Psi$.

The results are shown in Table~\ref{tab:ablation}. We can see that all of these components are quite important. Without Dropout \cite{srivastava2014dropout}, the architecture cannot find \emph{storage} positions to place an object if the goal is blocked, making it challenging to swap two objects -- a deterministic network could potentially learn this behavior, but it would be a much more difficult and less generalizable solution, and perhaps harder to learn from random imitation data.
Without $r_\Psi$, we see much longer plans as the model makes numerous small adjustments without terminating, as evidenced by the lower final error.
Without object selection, we select random objects that are not at their final positions, also leading to more inefficiency.



\subsection{Real Robot Experiments}
Finally, we performed a set of real-world experiments using a Franka Panda robot arm with an externally mounted Intel Realsense L515 camera. We set up the two types of testing scenarios i) swapping objects to achieve a given target arrangement and ii) sorting objects into different categories similar to the given target observation. Fig.~\ref{fig:cover} shows the robot sorting bowls and mugs, Fig.~\ref{fig:swap1} shows the swapping of two objects and  Fig.~\ref{fig:swap2} demonstrates the swapping tasks based on NeRP's predicted actions. However, the major limiting factors in real-robot tasks were noisy scene segmentation and feature extractions, which often led to rearrangement failure due to incorrect object correspondences, as also highlighted in the supplementary video.  

We see in all these different planning setups that despite being trained on synthetic data with only five object categories, NeRP generalizes to novel problem settings with high performance. In sim-to-real transfer, it generates actions for the low-level controller to move the object, while in other scenarios, the objects were directly teleported to the best placement location. 
See the supplementary materials for additional examples of NeRP real-world execution.


\section{Conclusions}
We presented Neural Rearrangement Planning (NeRP), a deep neural network-based rearrangement approach for unknown objects. NeRP can rearrange unseen objects without models, and works in an end-to-end fashion, given segmented point-clouds from an RGB-D camera. We evaluate NeRP on challenging problems and demonstrate its sim-to-real generalizations. NeRP relies on scene segmentation and correspondence matching techniques to generate a scene graph between the current and target observations. The scene graph is used to generate a sequence of intermediate object selection and their placement actions for reaching the given target arrangement. However, in our sim-to-real transfer experiments, we observed, as also shown in our supplementary video, that existing scene segmentation and feature-based object correspondence techniques often fail in the real-setup. This results in an incorrect scene graph and, therefore NeRP behavior. 

Hence, in our future works, we plan to augment NeRP with robust segmentation and feature matching algorithms to enhance its real-world settings' performance. Another of our future objectives is to extend NeRP for model-free planning in SE-3 space to compute both relative translation and rotations for transforming point sets into complex shapes and arrangements. 

\bibliographystyle{plainnat}
\bibliography{references}

\end{document}